\documentclass[pdflatex,sn-mathphys-num,iicol]{sn-jnl} 

\usepackage{graphicx}%

\usepackage{subcaption}
\usepackage{tikz}
\usetikzlibrary{arrows.meta,positioning,shapes.geometric}
\usepackage{multirow}%
\usepackage{amsmath,amssymb,amsfonts}%
\usepackage{amsthm}%
\usepackage{mathrsfs}%
\usepackage[title]{appendix}%
\usepackage{xcolor}%
\usepackage{textcomp}%
\usepackage{manyfoot}%
\usepackage{booktabs}%
\usepackage{algorithm}%
\usepackage{algorithmicx}%
\usepackage{algpseudocode}%
\usepackage{listings}%

\theoremstyle{thmstyleone}%
\theoremstyle{thmstyletwo}%

\theoremstyle{thmstylethree}%
\newtheorem{definition}{Definition}%

\begin{document}

\title{GPU-Accelerated Deep Learning for Heatwave Prediction and Urban Heat Risk Assessment}

\author*[1]{\fnm{Adis} \sur{Alihodzic}~(ORCID: 0000-0003-0761-1667)}\email{adis.alihodzic@pmf.unsa.ba}

\affil*[1]{\orgdiv{Department of Mathematical and Computer Sciences},
	\orgname{Faculty of Science, University of Sarajevo},
	\orgaddress{\city{Sarajevo}, \country{Bosnia and Herzegovina}}}	
			
\abstract{Heatwaves are an important problem in cities, and climate change makes this problem more difficult. In this paper, we present a GPU-based deep learning framework for next-day prediction of urban thermal conditions and for heat risk assessment. The study was carried out in Sarajevo by using MODIS land surface temperature data and Open-Meteo forecast data. We tested several models, including convolutional models and spatiotemporal models. Among them, ConvLSTM with a mixed loss function gave the best results. The obtained values were MAE = 0.2293, RMSE = 0.3089, and $R^2 = 0.8877$. The experiments also showed that results can be improved by using longer temporal series and additional meteorological variables. Since the framework was implemented on a GPU and trained with mixed precision, the execution time was reduced. Based on the predicted temperature fields, it was also possible to combine hazard information with exposure and vulnerability data in order to generate city heat risk maps. The proposed framework can be used as a practical basis for city heat analysis.}


\maketitle

\section{Introduction}\label{sec1}

Heatwaves are the most effective climate extreme today. It is well known that they have become more frequent, more severe, and longer in many regions. Their effects can be seen in public health, energy demand, labor productivity, and urban infrastructure. Recent studies have revealed that heat-related mortality is strongly associated with extreme hot periods, especially in cities, where the urban environment can further increase thermal stress \cite{CHEVAL2024100603,CUERDOVILCHES2023164412,Huang2023}. Also, urban areas are particularly sensitive to extreme heat, as dense construction, impervious materials, reduced vegetation, and limited air circulation often make cities warmer than surrounding rural areas. This phenomenon is known as the urban heat island. During heatwaves, the combined effects of climate change and urban heat island effects may aggravate thermal conditions. Therefore, predicting urban heat conditions and estimating urban heat risk have become essential topics in environmental modeling, public health, and urban planning \cite{CHEVAL2024100603,CUERDOVILCHES2023164412,Hsu2021}. Another crucial point is that urban heat does not affect all elements of a city equally. Some regions of the city have more vegetation and better ventilation, while others are represented by dense built-up areas and large impervious surfaces. Population exposure is also not uniform across all locations. Recent studies have shown that urban heat exposure may be spatially uneven and closely linked to demographic and socioeconomic patterns. Because of this, urban heat analysis should not be based solely on temperature predictions but should also incorporate exposure and vulnerability factors \cite{Hsu2021,Huang2023,Pan2024,DAmbrosio2023}. Satellite products provide spatially detailed land surface temperature information, while gridded population datasets can be used to estimate population exposure \cite{YE2025100870,Tatem2017}. MODIS land surface temperature products are widely used in thermal remote sensing and have been extensively validated in the literature \cite{WAN200859,DUAN201916}. Population data from the WorldPop project are also often used in spatial exposure analysis \cite{Tatem2017,Lloyd2017}. Conventional heatwave analysis has commonly relied on statistical methods, physical modeling, and conventional machine learning approaches \cite{Geophysics2023,su17083747}. Although these methods are appropriate, they may be limited when the process exhibits strong spatial and temporal dependencies simultaneously. For that reason, deep learning methods have received increasing attention in recent years because they have shown good results in spatiotemporal prediction problems \cite{BOUDREAULT2025109965,atmos16010082}. Convolutional neural networks are suitable for learning spatial patterns \cite{10.1007/978-981-19-1122-4-47}, while ConvLSTM architectures are designed to model spatial and temporal structure jointly. Besides prediction accuracy, computational efficiency is also important in practice. High-resolution urban heat modeling involves large spatiotemporal datasets and repeated model training, which may require substantial computational time \cite{ijgi4042306,hydrology11080127}. Therefore, GPU acceleration plays an important practical role, enabling faster training and inference in modern deep learning workflows \cite{Pandey2022,s24020514}. This is especially useful when the forecast results need to be transformed into urban heat risk layers for practical decision support \cite{LiWang2021,Pan2024}. In this chapter, heatwave prediction and urban heat risk assessment are considered from a practical perspective. The main idea is to combine satellite-derived land-surface temperatures with daily meteorological forcing within a GPU-accelerated deep learning framework. Two model families are considered, namely a CNN baseline and a ConvLSTM model. In addition, a simplified urban heat risk layer is constructed by combining predicted thermal intensity with exposure and vulnerability information. In that way, this chapter connects temperature prediction with a risk-based interpretation that can be useful in climate adaptation and urban planning studies \cite{Pan2024,DAmbrosio2023}. The main contributions of this chapter are as follows. First, a practical framework for combining satellite-derived thermal data and daily meteorological forcing is presented for short-term urban heat prediction. Second, the role of GPU acceleration in training and evaluation of deep learning models is discussed. Third, the influence of dataset design, temporal coverage, and multi-location meteorological forcing on predictive performance is experimentally analyzed. Fourth, a simplified methodology for generating urban heat risk maps from predicted thermal conditions, exposure factors, and vulnerability factors is outlined. The remainder of this chapter is organized as follows. Section~2 presents related work on heatwave prediction, urban heat analysis, and deep learning approaches. Section~3 describes the study area, data sources, and preprocessing steps. Section~4 introduces the proposed GPU-accelerated framework and the prediction models considered. Section~5 explains the construction of the urban heat risk layer. Section~6 presents the experimental results and performance analysis. Finally, Section~7 gives the concluding remarks and possible directions for future work.

\section{Related Work}\label{sec2}

\subsection{Deep learning for heatwave and urban thermal prediction}\label{subsec1}

Recent studies have shown growing interest in deep learning for heat-related prediction tasks. In the broader area of extreme heat forecasting, deep learning models have already been applied to predict extreme heat events from meteorological variables, and the reported results indicate that such data-driven approaches can achieve good predictive performance \cite{Shafiq2025}. Deep learning has also been used in urban thermal studies, especially for land surface temperature prediction and urban heat island analysis, including frameworks based on multisensor data and machine learning methods for prediction and assessment \cite{10938603,Wang2025}. However, there are still relatively few studies that combine satellite-derived thermal fields with daily meteorological forcing within a practical framework for next-day urban thermal prediction.

\subsection{Urban heat risk assessment and vulnerability mapping}\label{subsec2}

Another important direction is the transformation of thermal information into risk-oriented spatial analysis. Recent studies demonstrate that urban heat assessment should not be limited to temperature mapping but should also incorporate exposure, vulnerability, and population-related factors \cite{rs16163032}. In this sense, GIS- and remote-sensing-based frameworks are often used to combine thermal indicators with environmental and demographic layers to cause urban heat vulnerability or risk maps \cite{Pan2024,DAmbrosio2023}. However, many of these studies are mostly descriptive or retrospective, while there are still fewer practical strategies that directly connect short-term thermal prediction with simplified urban heat risk mapping.

\subsection{GPU acceleration in environmental deep learning}\label{subsec3}

The third important direction is computational efficiency. Deep learning workflows in environmental applications usually incorporate large spatiotemporal datasets, repeated model training, and computationally demanding mapping procedures, which makes acceleration important in practice \cite{Pandey2022}. In urban heat applications, GPU-based computation has already been utilized to accelerate heat-exposure mapping, and some studies documented reductions in computation time of more than 99\% for specific spatial estimations \cite{LiWang2021}. In a broader sense, fast deduction is also essential for near-real-time prediction and repeated evaluation \cite{Kalfarisi2022}. However, GPU acceleration is still often treated only as an implementation detail in the background, and less often as an explicit part of an integrated framework for urban thermal prediction and risk-oriented mapping. In general, the literature suggests that deep learning-based heat prediction \cite{Ge2025,Lyu2022}, urban heat risk assessment \cite{atmos14020343}, and GPU-accelerated computation \cite{Kalfarisi2022} have mostly been examined separately.

\section{Study Area, Data Sources, and Preprocessing}\label{sec3}

\subsection{Study area: Sarajevo}\label{subsec4}

\begin{figure}[!t]
	\centering
	\begin{subfigure}[t]{0.4\textwidth}
		\centering
		\includegraphics[width=\textwidth]{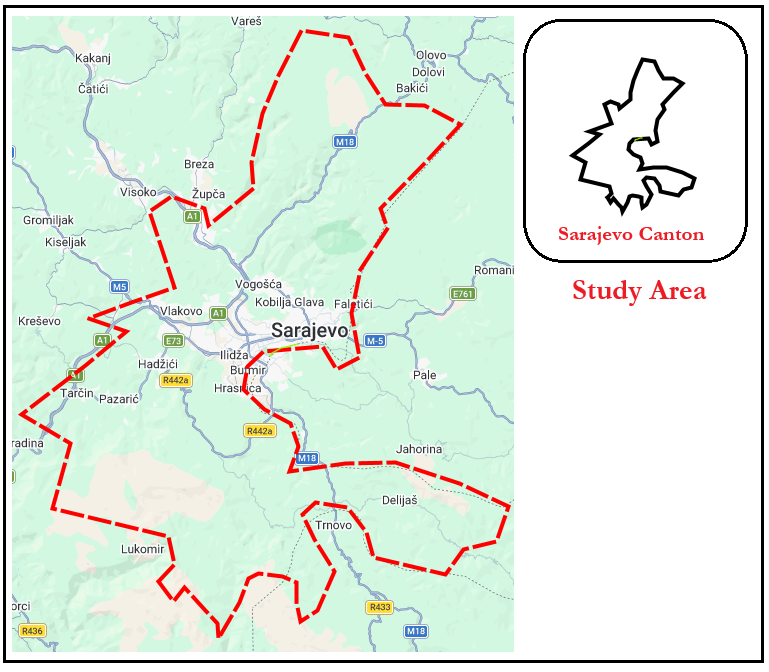}
		\caption{Study area: Sarajevo Canton.}
		\label{fig:sarajevo_study_area}
	\end{subfigure}
	\hfill
	\begin{subfigure}[t]{0.48\textwidth}
		\centering
		\includegraphics[width=0.85\textwidth]{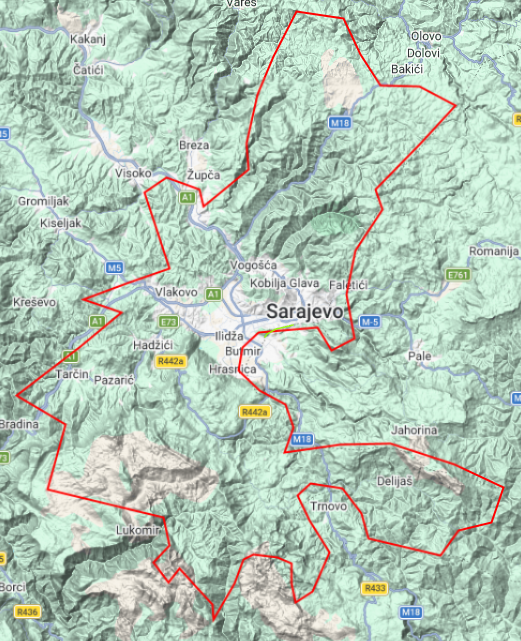}
		\caption{Terrain-based spatial context of Sarajevo.}
		\label{fig:sarajevo_topography}
	\end{subfigure}
	\caption{Spatial context of the study area. Panel (a) shows the geographic extent of Sarajevo used in this chapter, while panel (b) illustrates the surrounding topographic structure that contributes to spatial variability in urban thermal behavior.}
	\label{fig:study_area_context}
\end{figure}

The case study for urban heat analysis encloses Sarajevo and its surrounding neighborhoods, which were picked because their thermal behavior is influenced by a dense urban structure, valley topography, heterogeneous land cover, and ongoing urban development. In Fig.~\ref{fig:study_area_context}(a), the studied area combines urban zones, residential districts with distinct construction densities, central transport corridors, and encircling peri-urban space. Sarajevo and its surrounding urban areas are suitable for studying heat-related phenomena, as summer heat may be further intensified by limited air circulation and local atmospheric conditions. This is also endorsed by the terrain context presented in Fig.~\ref{fig:study_area_context}(b), where the city is encountered within a valley system and influenced by the surrounding elevated terrain. Such a spatial configuration makes Sarajevo a useful example for studying the relationship between broader meteorological forcing and local urban thermal response.

\subsection{Satellite-derived thermal data: MODIS LST}\label{subsec6}

\begin{figure}[!t]
	\centering
	\begin{subfigure}[t]{0.48\textwidth}
		\centering
		\includegraphics[width=0.9\textwidth]{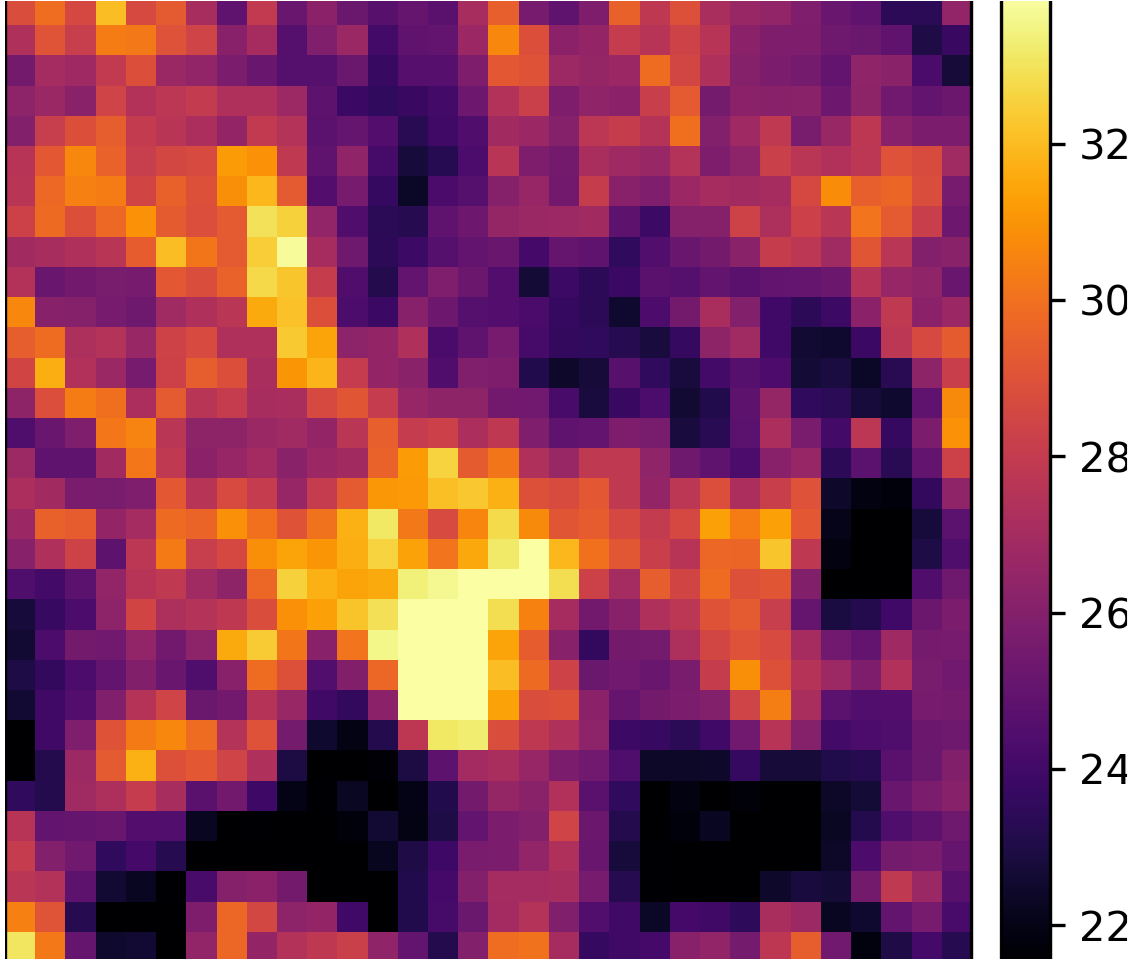}
		\caption{MODIS LST field over Sarajevo on 2022-07-14.}
		\label{fig:modis_lst_32x32_a}
	\end{subfigure}
	\hfill
	\begin{subfigure}[t]{0.48\textwidth}
		\centering
		\includegraphics[width=0.9\textwidth]{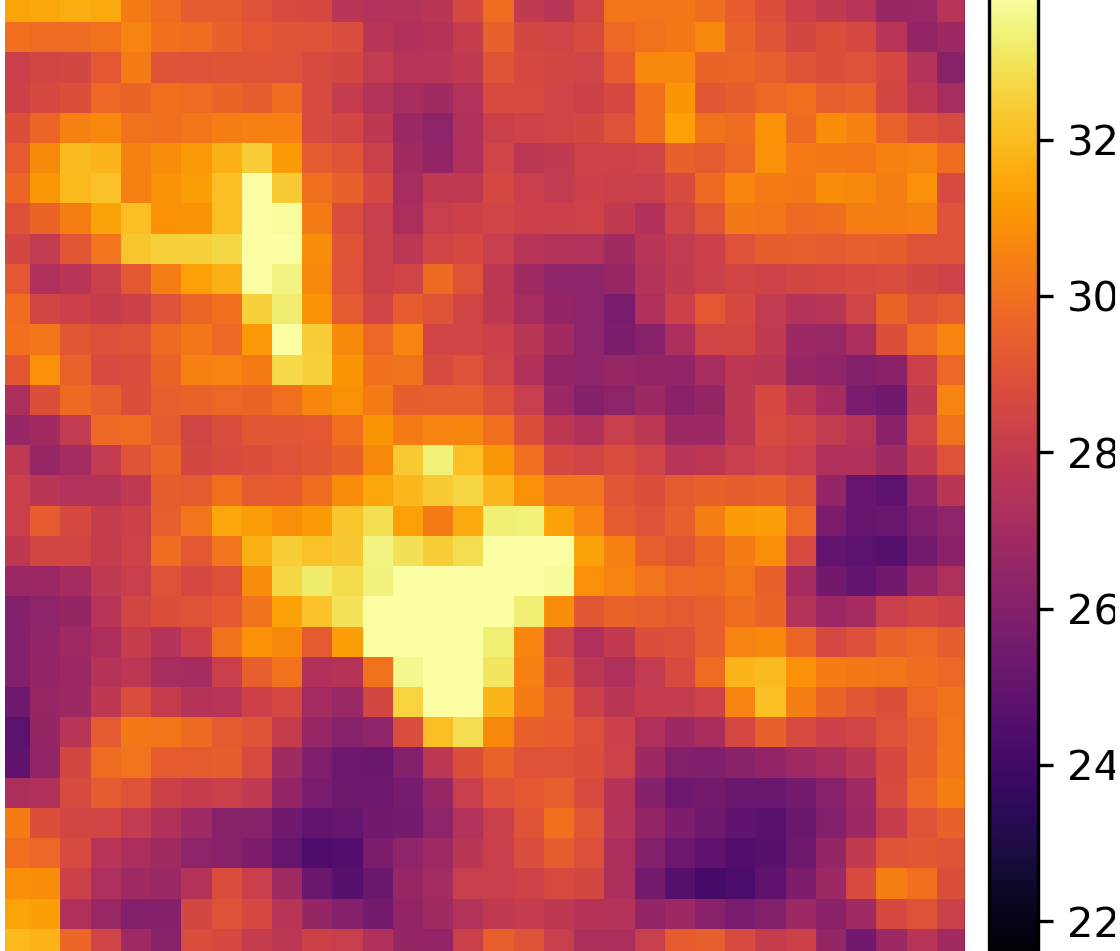}
		\caption{MODIS LST field over Sarajevo on 2017-07-31.}
		\label{fig:modis_lst_32x32_b}
	\end{subfigure}
	\caption{Examples of MODIS-derived summer land surface temperature fields over Sarajevo. The panels show representative daily thermal maps on a $32 \times 32$ grid, highlighting the spatial variability of urban surface temperature patterns across different dates.}
	\label{fig:modis_lst_example}
\end{figure}

The main thermal target considered in this paper is land surface temperature obtained from the MODIS sensor. MODIS LST products are widely used in urban climate and environmental studies because they provide spatially explicit thermal observations with regular temporal coverage. In this study, MODIS-derived thermal maps were used to represent day-to-day urban thermal conditions over Sarajevo. Some representative examples of summer thermal fields extracted from the prepared dataset are shown in Fig.~\ref{fig:modis_lst_example}, where the spatial temperature variability over the study area is clearly visible across different dates. Each valid MODIS scene was transformed into a spatial temperature field over Sarajevo and stored in a standardized form suitable for predictive modeling. After cropping to the study area and removing invalid observations where necessary, all target fields were arranged on a common $32 \times 32$ grid. In this way, a compact but still spatially informative thermal representation was obtained, suitable for deep learning and able to preserve the main spatial differences important for next-day urban thermal prediction.

\subsection{Meteorological forcing data: Open-Meteo}\label{subsec5}

\begin{figure}[!t]
	\centering
	\includegraphics[width=0.5\textwidth]{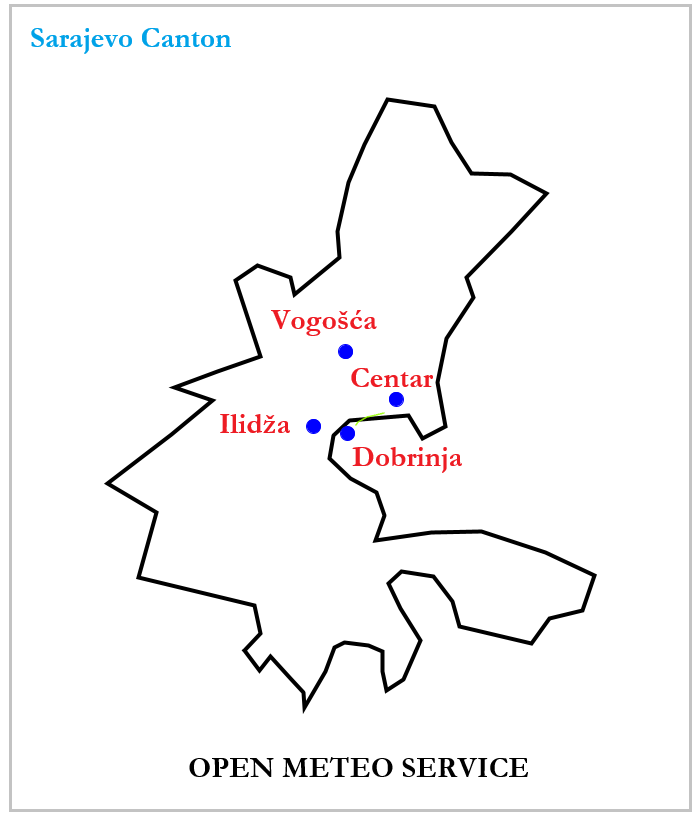}
	\caption{Representative spatial distribution of the Sarajevo locations used to obtain daily meteorological forcing variables from the Open-Meteo service. Using multiple locations provides a more representative characterization of atmospheric conditions across the study area than relying on a single-point record.}
	\label{fig:openmeteo_points}
\end{figure}

In addition to thermal imagery, the proposed predictive framework uses daily meteorological forcing variables from the Open-Meteo service. These variables were included because urban thermal behavior depends not only on land-surface conditions but also on surrounding atmospheric conditions. To obtain a more representative description of daily forcing over the city, meteorological data were collected from several locations in Sarajevo rather than using a single point. Their spatial distribution is illustrated in Fig.~\ref{fig:openmeteo_points}. The last set of daily variables possesses temperature-related indicators, humidity, precipitation, radiation, cloudiness, and wind characteristics. Hence, the predictive framework merges spatial thermal information from satellite measerments with time-aligned daily meteorological forcing. The usage of multiple sampling spots was vital, as Sarajevo has spatially distinct urban and topographic markers, and a single meteorological record could not properly capture the variability in causing conditions across the more comprehensive study area.

\subsection{Data preprocessing and dataset construction}\label{subsec8}

\begin{figure}[t]
	\centering
	\resizebox{\columnwidth}{!}{%
		\begin{tikzpicture}[
			font=\small,
			>=stealth,
			box/.style={
				draw=black!70,
				rounded corners=6pt,
				align=center,
				minimum height=1.0cm,
				text width=3.2cm,
				inner sep=4pt
			},
			in1/.style={box, fill=red!12},
			in2/.style={box, fill=blue!12},
			in3/.style={box, fill=green!12},
			proc/.style={box, fill=gray!10, text width=3.8cm},
			out1/.style={box, fill=yellow!25},
			arrow/.style={->, thick}
			]
			
			\node[font=\bfseries\small] at (0,5.6) {Input data};
			\node[font=\bfseries\small] at (5.5,5.6) {Preprocessing};
			\node[font=\bfseries\small] at (10.6,5.6) {Model-ready outputs};
			
			\node[in1] (modis) at (0,4.2) {MODIS LST\\scenes};
			\node[in2] (meteo) at (0,2.6) {Open-Meteo daily\\variables};
			\node[in3] (area)  at (0,1.0) {Study area\\definition};
			
			\node[proc] (crop)   at (5.5,4.8) {Spatial cropping\\to Sarajevo};
			\node[proc] (filter) at (5.5,3.5) {Invalid-pixel\\filtering};
			\node[proc] (align)  at (5.5,2.2) {Temporal alignment of\\thermal and meteorological data};
			\node[proc] (norm)   at (5.5,0.9) {Normalization / scaling};
			\node[proc] (seq)    at (5.5,-0.4) {Sequence construction\\for temporal models};
			
			\node[out1] (cnn)   at (10.6,4.0) {CNN input};
			\node[out1] (clstm) at (10.6,2.2) {ConvLSTM input\\sequences};
			\node[out1] (target)at (10.6,0.2) {Next-day $32\times 32$\\thermal target};
			
			\draw[arrow] (modis.east) -- (crop.west);
			\draw[arrow] (meteo.east) -- (align.west);
			\draw[arrow] (area.east) -- (crop.west);
			
			\draw[arrow] (crop.south) -- (filter.north);
			\draw[arrow] (filter.south) -- (align.north);
			\draw[arrow] (align.south) -- (norm.north);
			\draw[arrow] (norm.south) -- (seq.north);
			
			\draw[arrow] (seq.east) -- (cnn.west);
			\draw[arrow] (seq.east) -- (clstm.west);
			\draw[arrow] (seq.east) -- (target.west);
			
		\end{tikzpicture}%
	}
	\caption{Preprocessing pipeline from raw inputs to model-ready outputs.}
	\label{fig:preprocessing_pipeline}
\end{figure}
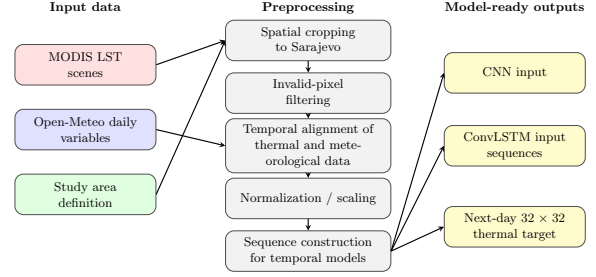

A preprocessing stage was necessary in order to transform the heterogeneous raw data into a consistent dataset suitable for deep learning. Since the considered data sources differ in spatial resolution, temporal frequency, and format, all inputs had to be organized within a common spatial and temporal framework. The overall workflow is summarized in Fig.~\ref{fig:preprocessing_pipeline}, which shows the main steps from raw data acquisition to the construction of samples ready for model training. For the thermal component, invalid MODIS observations were removed as needed, and the remaining scenes were converted into standardized daily raster fields over Sarajevo. For the meteorological component, daily forcing variables were synchronized with the dates of the available thermal observations. This temporal alignment step was important because each training sample had to connect predictor variables from a given temporal window with the target thermal field of the next day. After synchronization, the dataset was divided into input-output pairs for supervised learning. In the case of sequence-based models such as ConvLSTM, the data were further arranged into short temporal sequences so that the network could learn both spatial structure and temporal evolution. Additional preprocessing steps, such as normalization and scaling, were also applied to improve stability during training and numerical stability. As shown in Fig.~\ref{fig:preprocessing_pipeline}, the final result of preprocessing was not only a collection of aligned rasters and tabular variables, but a structured dataset prepared for the considered predictive architectures. The resulting dataset construction pipeline provided the basis for the experiments presented later in this chapter.

\section{Proposed GPU-Accelerated Framework}\label{sec4}

\subsection{Problem formulation}\label{subsec9}

The objective of the proposed framework is to predict urban thermal conditions for the next day over the Sarajevo study area and to provide outputs that can later be used for simplified urban heat risk assessment. Let $X_t$ present information at day $t$, including the recent thermal state of the city represented by spatial temperature fields and the corresponding meteorological forcing variables. The purpose of the forecast is to comprehend a mapping such as
\[
f: X_t \mapsto \hat{Y}_{t+1},
\]
where $\hat{Y}_{t+1}$ is the prediction for day $t+1$. In other words, the model output is a raster-like temperature map representing the expected spatial distribution of thermal intensity across the study area for the following day. Depending on the considered architecture, the input may be given either as a channel-stacked spatial representation or as a short temporal sequence of previous observations. In all cases, the goal is not only to predict a single scalar temperature value but to reconstruct the complete spatial thermal pattern, which makes the problem inherently spatiotemporal. To achieve the same forecast goal, two model families were assessed: the first is a CNN-based model used as a baseline predictor of spatial thermal patterns, while the second is a ConvLSTM model made to capture both spatial and temporal dependencies. The CNN and ConvLSTM models were trained on the same multi-source dataset and evaluated by metrics such as mean absolute error (MAE), root mean square error (RMSE), and the coefficient of determination ($R^2$).

\subsection{Prediction models}\label{subsec10}

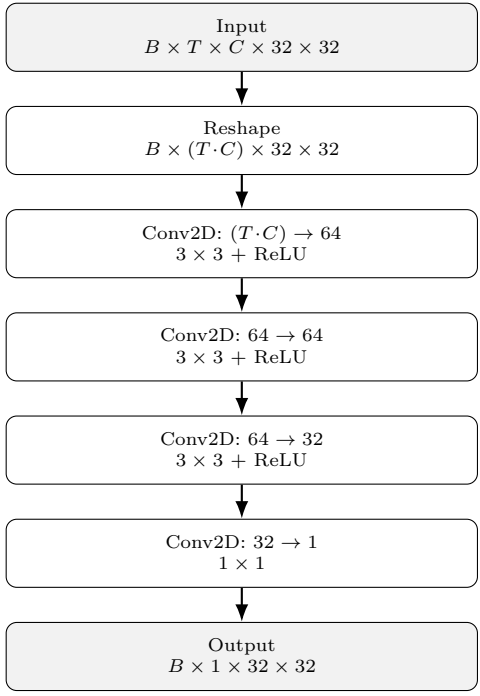
\begin{figure}[!t]
	\centering
	\begin{tikzpicture}[
		node distance=0.45cm,
		>=Latex,
		box/.style={
			draw,
			rounded corners,
			align=center,
			minimum width=0.82\columnwidth,
			minimum height=0.9cm,
			font=\footnotesize,
			inner sep=3pt
		},
		io/.style={
			draw,
			rounded corners,
			fill=gray!10,
			align=center,
			minimum width=0.82\columnwidth,
			minimum height=0.9cm,
			font=\footnotesize,
			inner sep=3pt
		}
		]
		
		\node[io] (in) {Input\\$B \times T \times C \times 32 \times 32$};
		\node[box, below=of in] (reshape) {Reshape\\$B \times (T\!\cdot\!C) \times 32 \times 32$};
		\node[box, below=of reshape] (c1) {Conv2D: $(T\!\cdot\!C)\rightarrow 64$\\$3\times3$ + ReLU};
		\node[box, below=of c1] (c2) {Conv2D: $64\rightarrow 64$\\$3\times3$ + ReLU};
		\node[box, below=of c2] (c3) {Conv2D: $64\rightarrow 32$\\$3\times3$ + ReLU};
		\node[box, below=of c3] (c4) {Conv2D: $32\rightarrow 1$\\$1\times1$};
		\node[io, below=of c4] (out) {Output\\$B \times 1 \times 32 \times 32$};
		
		\draw[->, thick] (in) -- (reshape);
		\draw[->, thick] (reshape) -- (c1);
		\draw[->, thick] (c1) -- (c2);
		\draw[->, thick] (c2) -- (c3);
		\draw[->, thick] (c3) -- (c4);
		\draw[->, thick] (c4) -- (out);
		
	\end{tikzpicture}
	\caption{CNN baseline architecture used for next-day urban thermal prediction.}
	\label{fig:cnn_architecture}
\end{figure}

\begin{figure}[!t]
	\centering
	\begin{tikzpicture}[
		node distance=0.45cm,
		>=Latex,
		box/.style={
			draw,
			rounded corners,
			align=center,
			minimum width=0.82\columnwidth,
			minimum height=0.9cm,
			font=\footnotesize,
			inner sep=3pt
		},
		io/.style={
			draw,
			rounded corners,
			fill=gray!10,
			align=center,
			minimum width=0.82\columnwidth,
			minimum height=0.9cm,
			font=\footnotesize,
			inner sep=3pt
		}
		]
		
		\node[io] (in) {Input sequence\\$B \times T \times C \times 32 \times 32$};
		\node[box, below=of in] (seq) {Temporal processing\\$x_1,\dots,x_T$};
		\node[box, below=of seq] (cell) {ConvLSTM cell\\input channels $=C$, hidden dim $=32$\\$3\times3$ gates};
		\node[box, below=of cell] (ht) {Final hidden state\\$B \times 32 \times 32 \times 32$};
		\node[box, below=of ht] (head1) {Conv2D: $32\rightarrow 32$\\$3\times3$ + ReLU};
		\node[box, below=of head1] (head2) {Conv2D: $32\rightarrow 1$\\$1\times1$};
		\node[io, below=of head2] (out) {Output\\$B \times 1 \times 32 \times 32$};
		
		\draw[->, thick] (in) -- (seq);
		\draw[->, thick] (seq) -- (cell);
		\draw[->, thick] (cell) -- (ht);
		\draw[->, thick] (ht) -- (head1);
		\draw[->, thick] (head1) -- (head2);
		\draw[->, thick] (head2) -- (out);
		
	\end{tikzpicture}
	\caption{ConvLSTM architecture used in the final experiments.}
	\label{fig:convlstm_architecture}
\end{figure}
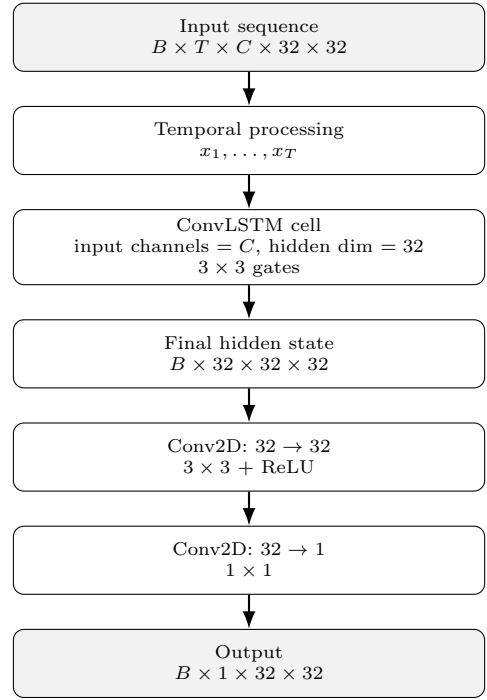

We analyze the CNN model in the most detail. We introduced this model in order to use a simple deep learning predictor for spatial thermal patterns. The main reason for using it is its compact, fully convolutional architecture, which can process raster-like inputs while remaining simple enough for stable training and direct implementation. We assume that the input tensor is of size $B \times T \times C \times H \times W$, where $B$, $T$, and $C$ are the batch size, sequence length, and the number of channels for one day, respectively. Also, inside the CNN model, the temporal and channel dimensions were merged before convolutional processing, so that the input is mapped into the shape of dimension $B \times (T \cdot C) \times H \times W$. Except for the mentioned,  the network includes three $3 \times 3$ convolutional layers with padding $1$, followed by ReLU activations, and a final $1 \times 1$ convolution, producing the predicted single-channel thermal map. For the hidden width equal to $64$, the channel progression is 

\[
(T \cdot C) \rightarrow 64 \rightarrow 64 \rightarrow 32 \rightarrow 1.
\]
Therefore, the CNN baseline follows the structure:
\begin{itemize}
	\item Conv2D$(T\!\cdot\!C, 64, 3 \times 3, \text{padding}=1)$ + ReLU,
	\item Conv2D$(64, 64, 3 \times 3, \text{padding}=1)$ + ReLU,
	\item Conv2D$(64, 32, 3 \times 3, \text{padding}=1)$ + ReLU,
	\item Conv2D$(32, 1, 1 \times 1)$.
\end{itemize}
Although the architecture lacks pooling layers, the spatial resolution is kept throughout the network, and the output is fixed to a $32 \times 32$ thermal matrix. However, temporal development is not explicitly modeled, except through the implicit information furnished by channel stacking. The main predictive model used in this chapter is a ConvLSTM architecture. Unlike the CNN baseline, this model keeps the temporal structure explicitly and processes the input in its original sequential form, that is, as a tensor of size $B \times T \times C \times H \times W$. At each time step, the current input and the previous hidden state are concatenated and passed through a $3 \times 3$ convolution with padding $1$. In that way, the input, forget, output, and candidate gates of the ConvLSTM cell are generated. In the final experiments, the hidden dimension was set to $32$. After the last time step, the final hidden state is passed to a convolutional prediction head. This head consists of one convolutional layer of size $3 \times 3$ with $32$ output channels, followed by a ReLU activation function, and a final convolutional layer of size $1 \times 1$ that produces the predicted single-channel thermal map. Therefore, the prediction head has the following form:
\begin{itemize}
	\item Conv2D$(32, 32, 3 \times 3, \text{padding}=1)$ + ReLU,
	\item Conv2D$(32, 1, 1 \times 1)$.
\end{itemize}
Such a design allows the ConvLSTM model to capture both the spatial arrangement of urban heat patterns and their short-term temporal evolution. This is important in the case of persistent heatwave conditions. The structure of the CNN baseline is shown in Fig.~\ref{fig:cnn_architecture}. The model first merges the temporal and channel dimensions, then applies a fully convolutional mapping to obtain the predicted thermal field for the next day. The ConvLSTM architecture used in the final experiments is shown in Fig.~\ref{fig:convlstm_architecture}. In contrast to the CNN baseline, it preserves the temporal structure and processes the input sequence recurrently before generating the final thermal map by a convolutional prediction head.

\subsection{GPU acceleration strategy}\label{subsec13}

Because the proposed framework involves repeated training and evaluation of multiple deep learning architectures on multi-source spatiotemporal data, computational efficiency becomes an important practical issue. For that reason, all experiments in this chapter are designed around GPU-accelerated execution. The experiments were carried out on a workstation equipped with an AMD64 processor, 16~GB of RAM, and Windows~11. Training was performed on CUDA-enabled hardware, and GPU acceleration was used in order to reduce computation time and allow repeated model evaluation under different dataset configurations. The implementation relies on CUDA-enabled tensor computation, which allows convolutions and recurrent updates to be executed in parallel. This substantially reduces the training time compared with a purely CPU-based workflow. In addition, mixed-precision training is employed where appropriate in order to further improve computational efficiency and reduce memory usage. This is particularly useful when working with image-like spatial data and sequence-based models, since both can lead to significant memory demands during training. From a methodological perspective, GPU acceleration is not treated only as a technical implementation detail, but as an integral part of the proposed practical framework. Without accelerated training, repeated experimentation with CNN and ConvLSTM models would be significantly more difficult, especially when multiple input configurations and loss formulations must be tested. The use of GPU resources therefore directly supports the experimental design of the chapter and helps make the proposed urban thermal prediction workflow computationally viable in practice.

\section{Urban Heat Risk Assessment Layer}\label{sec5}

Although the primary focus of this chapter is next-day urban thermal prediction, the obtained predictive maps can also be used as the basis for a simplified urban heat risk assessment framework. In climate-risk literature, risk is commonly interpreted through the interaction of hazard, exposure, and vulnerability, and this perspective has also been adopted in urban heat studies \cite{doi:10.1177/23998083241280746,doi:10.1371/journal.pone.0127277}. The main idea here is therefore to extend temperature prediction toward a more decision-oriented spatial interpretation by combining the predicted thermal field with additional exposure and vulnerability information. In this chapter, this risk-oriented extension is intentionally kept simple and transparent. Its purpose is not to provide a fully operational urban risk system, but rather to illustrate how GPU-accelerated thermal prediction can support practical urban heat analysis and spatial screening of potentially critical urban zones \cite{ZHANG2019852}.

\subsection{Hazard layer}\label{subsec14}

The hazard layer represents the spatial intensity of predicted thermal stress over the study area. In the proposed framework, it is derived directly from the predicted next-day thermal field produced by the deep learning models. Such an interpretation is consistent with earlier urban heat risk studies in which thermal indicators, including land surface temperature or related heat-stress measures, are used as the hazard component of the risk framework \cite{ZHANG2019852,doi:10.1177/23998083241280746}. Locations with higher predicted temperature values are interpreted as having higher thermal hazard, while locations with lower values correspond to lower hazard. For integration with the remaining components of the framework, the predicted field can be normalized to a common interval such as $[0,1]$, which is a standard step when combining heterogeneous indicators with different units and scales \cite{10.3389/fpubh.2022.989963,ZHANG2019852}. In this way, the hazard layer serves as the predictive component of the urban heat risk assessment, indicating where thermally critical conditions are expected to occur on the following day.

\subsection{Exposure layer}\label{subsec15}

The exposure layer is introduced to represent the spatial distribution of people or urban elements that may be affected by elevated thermal conditions. Its role is important because high thermal intensity alone does not necessarily indicate a high practical risk, unless the affected area also includes a considerable human presence or urban activity. In urban heat risk mapping, exposure is often represented by population density, pedestrian presence, residential concentration, or other indicators of human activity \cite{ZHANG2019852,doi:10.1177/23998083241280746}. In a simplified setting, the exposure layer can therefore be constructed from gridded demographic data or from proxy indicators such as residential density or built-up intensity. After alignment with the spatial grid of the hazard layer, the obtained surface is normalized so that higher values indicate stronger exposure \cite{10.3389/fpubh.2022.989963}. In this way, it becomes possible to distinguish between areas with low exposure to heat and those where heat may affect a larger number of people.

\subsection{Vulnerability layer}\label{subsec16}

The vulnerability layer describes the differential sensitivity of different parts of the city to thermal stress. In urban heat research, vulnerability is commonly associated with environmental, demographic, social, and urban-form characteristics that increase sensitivity or reduce coping capacity under hot conditions \cite{ijerph15040640,10.3389/fpubh.2022.989963,su15031820}. In real life, these elements may include limited vegetation, dense built-up structure, impervious surfaces, low ventilation potential, disadvantaged socioeconomic conditions, or the presence of particularly sensitive population groups. Several recent studies have shown that vulnerability-oriented heat assessments benefit from integrating both social indicators and spatial characteristics of the urban fabric \cite{ijerph15040640,su15031820}. In the present chapter, vulnerability is represented in a simplified manner through spatial indicators that align with the study grid. As in the case of hazard and exposure, the resulting layer is normalized before integration \cite{10.3389/fpubh.2022.989963}. Higher values indicate locations where predicted heat is expected to have stronger adverse implications due to less favorable local conditions.

\subsection{Risk index generation}\label{subsec17}

After the hazard, exposure, and vulnerability layers have been organized, they can be united into one simplified urban heat risk index. This follows the general idea of climate- and urban heat-risk assessment, where the most critical areas are those in which a strong thermal hazard co-occurs with high exposure and increased vulnerability \cite{10.3389/fpubh.2022.989963,ZHANG2019852}. One simple and transparent formulation is
\[
R(i,j) = H(i,j)\cdot E(i,j)\cdot V(i,j),
\]
where $H(i,j)$, $E(i,j)$, and $V(i,j)$ denote the normalized hazard, exposure, and vulnerability values at spatial position $(i,j)$, respectively. Multiplicative and similar composite formulations based on normalized indicators are often used in heat-vulnerability and heat-risk studies because they emphasize the joint effects of multiple factors, rather than relying solely on temperature \cite{10.3389/fpubh.2022.989963,ZHANG2019852}. The final risk surface can then be presented as a raster map or divided into ordinal categories such as low, moderate, high, and very high risk. Although this formulation is simplified, it is sufficient to show how next-day thermal forecasting can be extended towards interpretable urban heat risk assessment products \cite{10.3389/fpubh.2022.989963,doi:10.1177/23998083241280746}.

\section{Experimental Results}\label{subsec18}

The experimental part of this article was designed as a practical case study for next-day urban thermal prediction over Sarajevo. The goal was to predict the next daily land surface temperature (LST) map from a short temporal sequence of previous observations and meteorological forcing variables. All experiments were performed on CUDA-enabled hardware, and GPU acceleration was used during training in order to reduce training time and allow repeated model evaluation under different dataset configurations. The target output in all experiments was a normalized MODIS LST map of spatial size $32 \times 32$. Hence, each target map contains $1024$ pixel values. The input tensor for one sample had the general form
\[
X \in \mathbb{R}^{T \times C \times H \times W},
\]
where $T$ denotes the sequence length, $C$ denotes the number of channels per day, and $H=W=32$. In the final experiments, the sequence length was fixed to $T=3$. Two main dataset configurations were considered. The first configuration combined daily MODIS LST maps with daily meteorological forcing extracted from a single Open-Meteo location over Sarajevo. The second configuration extended both the temporal coverage and the number of meteorological inputs by using four representative Sarajevo locations together with a longer time interval. The final configuration of the best-performing ConvLSTM experiment is summarized in Table~\ref{tab:experimental_setup_short}. It includes the main characteristics of the input representation, the temporal setup, the training configuration, and the evaluation protocol used in the final stage of the study. All experiments were carried out on a workstation equipped with an AMD64 processor, 16~GB of RAM, and Windows~11.

\begin{table}[ht]
	\centering
	\caption{Final experimental setup used for the best-performing ConvLSTM model.}
	\label{tab:experimental_setup_short}
	\begin{tabular}{ll}
		\hline
		\textbf{Setting} & \textbf{Value} \\
		\hline
		Input data & MODIS LST \\
		& Open-Meteo daily variables \\
		Study period & 2010--2025 \\
		Meteorological locations & 4 \\
		Input channels per day & 41 \\
		Sequence length & 3 \\
		Input shape & $3 \times 41 \times 32 \times 32$ \\
		Output shape & $1 \times 32 \times 32$ \\
		Valid samples & 1871 \\
		Train/validation split & 80\% / 20\% \\
		Model & ConvLSTM \\
		Hidden dimension & 32 \\
		Optimizer & Adam \\
		Learning rate & 0.001 \\
		Batch size & 16 \\
		Epochs & 20 \\
		Loss & $0.7\,L_1 + 0.3\,L_2$ \\
		Training mode & GPU + mixed precision \\
		Metrics & MAE, RMSE, $R^2$ \\
		\hline
	\end{tabular}
\end{table}

\subsection{Dataset Construction}\label{subsec19}

The first dataset configuration was obtained by combining MODIS MOD11A1 daily daytime LST data with Open-Meteo daily meteorological variables for Sarajevo. In the case of one-location forcing, the following meteorological variables were used:
\begin{itemize}
	\item temperature\_2m\_mean,
	\item temperature\_2m\_max,
	\item temperature\_2m\_min,
	\item dew\_point\_2m\_mean,
	\item precipitation\_sum,
	\item wind\_speed\_10m\_mean,
	\item shortwave\_radiation\_sum.
\end{itemize}

These variables were combined with one MODIS LST channel, so that the total number of channels per day was equal to $8$. For the temporal sequence length equal to $3$, one input sample had the form
\[
3 \times 8 \times 32 \times 32.
\]
After preprocessing and removal of invalid cases, this baseline configuration contained $667$ valid samples. In the second configuration, the dataset was extended in several ways. First, the MODIS period was expanded to 2010--2025, while the initial configuration covered the period 2019--2024. Second, meteorological forcing was not collected from only one point, but from four Sarajevo locations: Sarajevo Center, Ilid\v{z}a, Dobrinja, and Vogo\v{s}\v{c}a. Third, the number of meteorological variables was increased to include:
\begin{itemize}
	\item temperature\_2m\_mean,
	\item temperature\_2m\_max,
	\item temperature\_2m\_min,
	\item dew\_point\_2m\_mean,
	\item precipitation\_sum,
	\item shortwave\_radiation\_sum,
	\item relative\_humidity\_2m\_mean,
	\item cloud\_cover\_mean,
	\item wind\_direction\_10m\_dominant,
	\item wind\_speed\_10m\_mean.
\end{itemize}

Since these variables were taken for four locations, the meteorological forcing contributed $40$ channels per day. Together with the MODIS LST channel, the total number of channels per day was $41$. Therefore, one input sample had the form
\[
3 \times 41 \times 32 \times 32.
\]
After alignment in time and removal of invalid cases, the final extended dataset contained $1871$ valid samples.

\subsection{Models and Training Procedure}\label{subsec20}

Two deep learning architectures were considered during the experiments: a CNN baseline and a ConvLSTM model. The CNN was used as a simpler reference model, while the ConvLSTM was expected to better capture spatiotemporal dependencies. In the final stage of the experiments, the ConvLSTM model clearly showed better behaviour and was therefore selected as the main model for detailed analysis. The training and validation sets were obtained using an $80\%/20\%$ split. Thus, for the extended dataset configuration, the model was trained on $1538$ samples and validated on $333$ samples. In all experiments, training was performed on GPU. Mixed precision was also used during training in order to improve computational efficiency. At first, a standard mean squared error loss was used. However, the predicted maps were often too smooth, even when the quantitative metrics were acceptable. For this reason, a hybrid loss was introduced:
\[
\mathcal{L} = \alpha \mathcal{L}_{1} + (1-\alpha)\mathcal{L}_{2},
\]
where $\mathcal{L}_{1}$ denotes the mean absolute error loss, $\mathcal{L}_{2}$ denotes the mean squared error loss, and $\alpha=0.7$ in our experiments. This modification led to both better numerical results and visually more realistic thermal maps.

\subsection{Evaluation Metrics}\label{subsec21}

The models were evaluated using three standard regression metrics:
\begin{itemize}
	\item mean absolute error (MAE),
	\item root mean squared error (RMSE),
	\item coefficient of determination ($R^2$).
\end{itemize}

These metrics were computed over the predicted and reference LST maps.
\subsection{Quantitative Results}

The most important quantitative results are summarized in Table~\ref{tab:main_results}. The first strong result was obtained on the MODIS + Open-Meteo single-location dataset 1 with $667$ samples. Using the ConvLSTM model with the hybrid loss, the method achieved the following values: $\textbf{MAE}=0.2543$, $\textbf{RMSE}=0.3349$, and $\textbf{R}^2=0.8606$. After extending the temporal range and replacing the single-location forcing with a multi-location meteorological representation, the results improved further. On the final dataset 2 with $1871$ samples, the ConvLSTM model achieved the following values: $\textbf{MAE}=0.2293$, $\textbf{RMSE}=0.3089$, and $\textbf{R}^2=0.8877$. These results show that both a stronger dataset and a richer meteorological representation improve next-day urban thermal prediction.

\begin{table}[ht]
	\centering
	\caption{Main quantitative results obtained with the ConvLSTM model.}
	\label{tab:main_results}
	\begin{tabular}{lcccc}
		\hline
		Dataset & Num. of & MAE & RMSE & $R^2$ \\
		configuration & samples &  &  & \\ \hline
		Dataset 1 & 667  & 0.2543 & 0.3349 & 0.8606 \\
		Dataset 2 & 1871 & 0.2293 & 0.3089 & 0.8877 \\
		\hline
	\end{tabular}
\end{table}

\begin{figure*}[h]
	\begin{center}$
		\begin{array}{ccc}
			\subfloat[First reference map.]{\includegraphics[scale=.2]{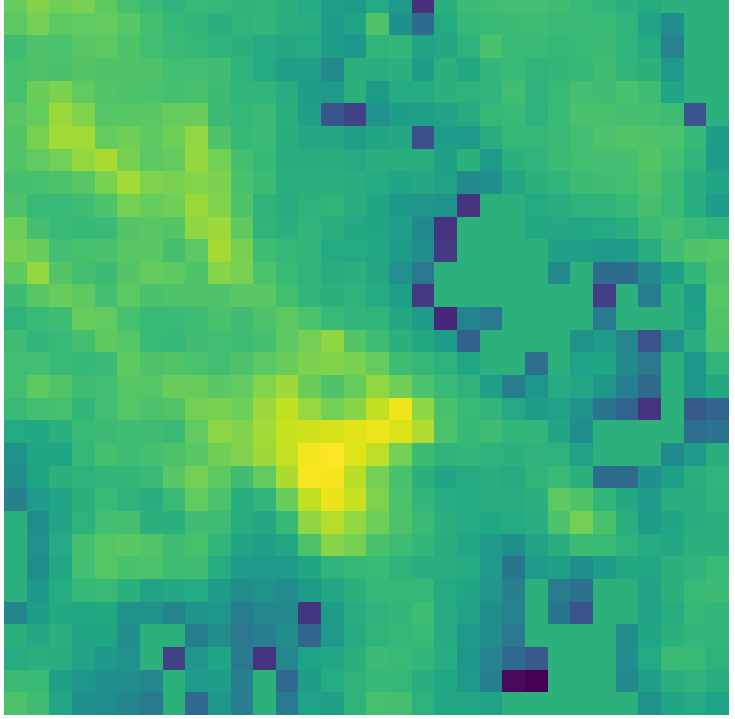}}&
			\subfloat[Second reference map.]{\includegraphics[scale=.2]{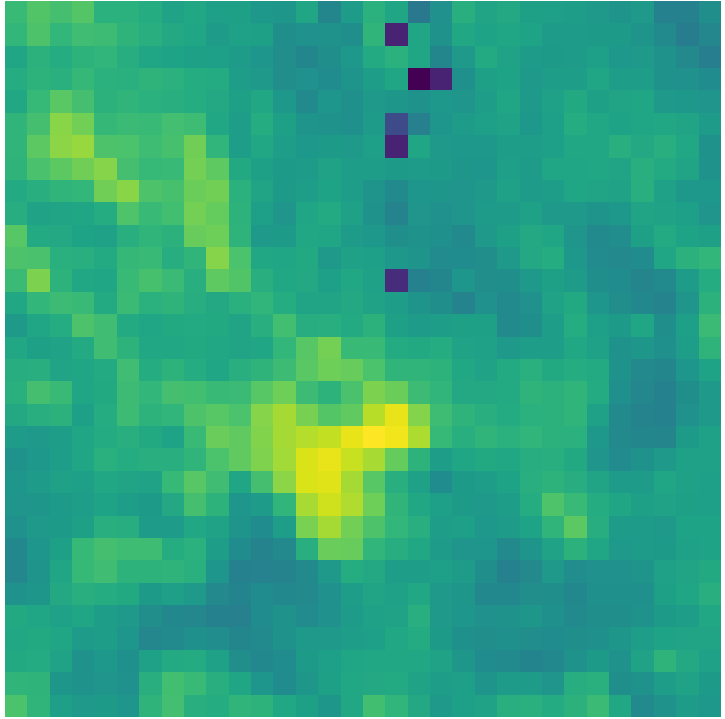}} &
			\subfloat[Third reference map.]{\includegraphics[scale=.2]{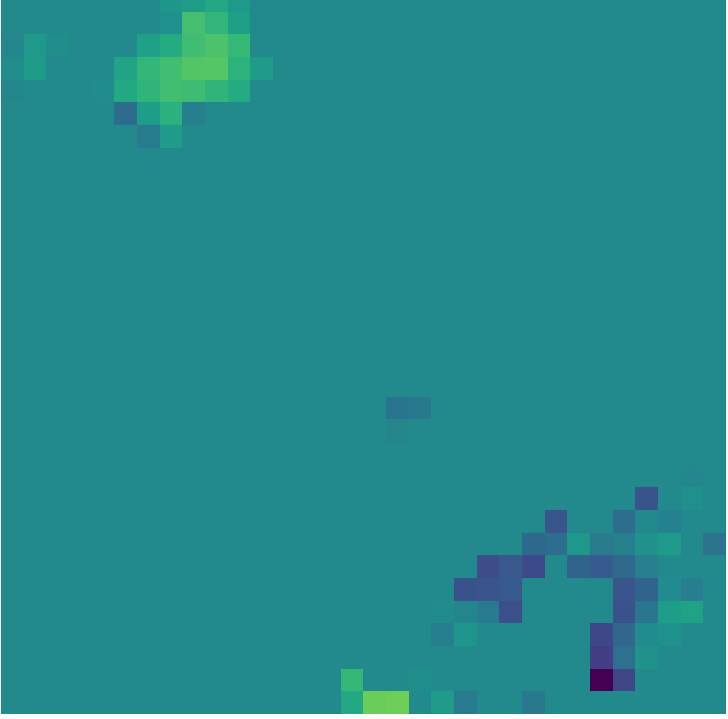}} \\
			\subfloat[First predicted map.]{\includegraphics[scale=.2]{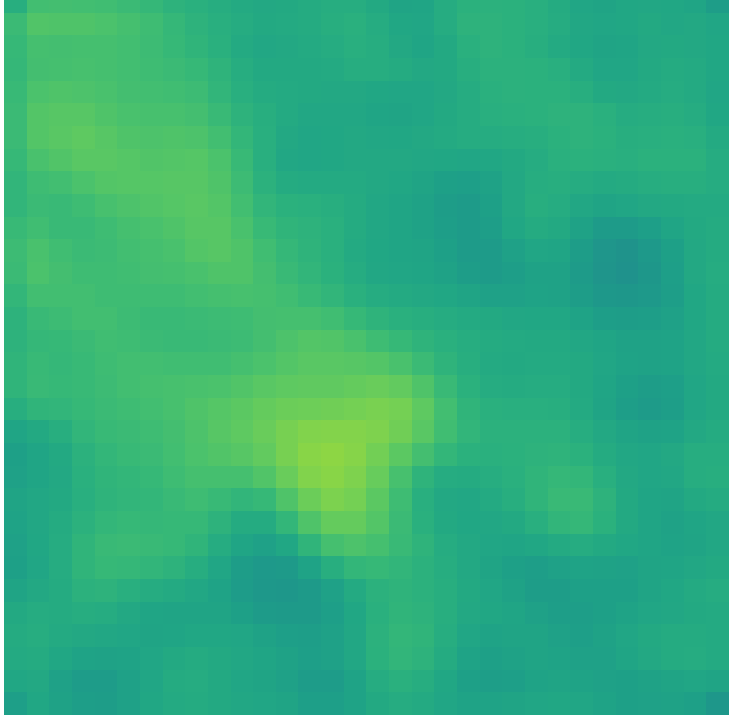}} &
			\subfloat[Second predicted map.]{\includegraphics[scale=.2]{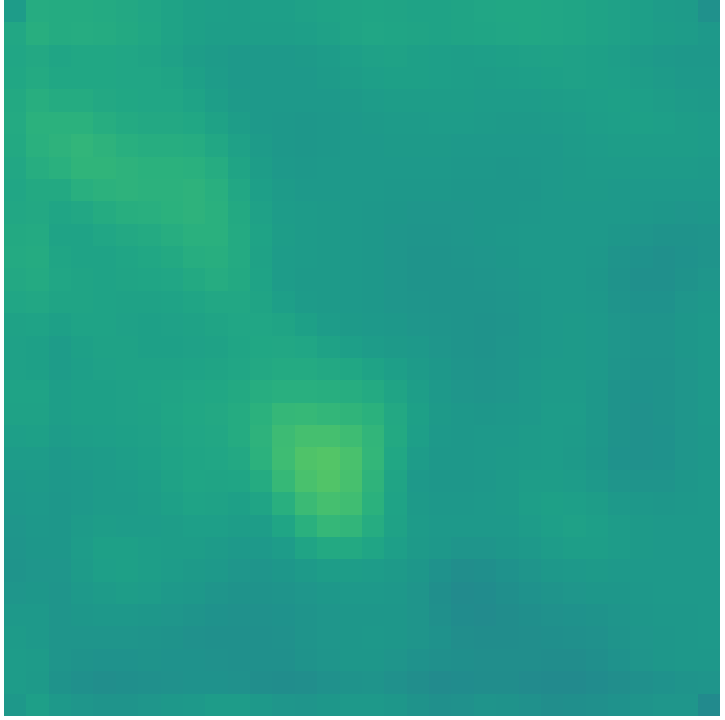}} &
			\subfloat[Third predicted map.]{\includegraphics[scale=.2]{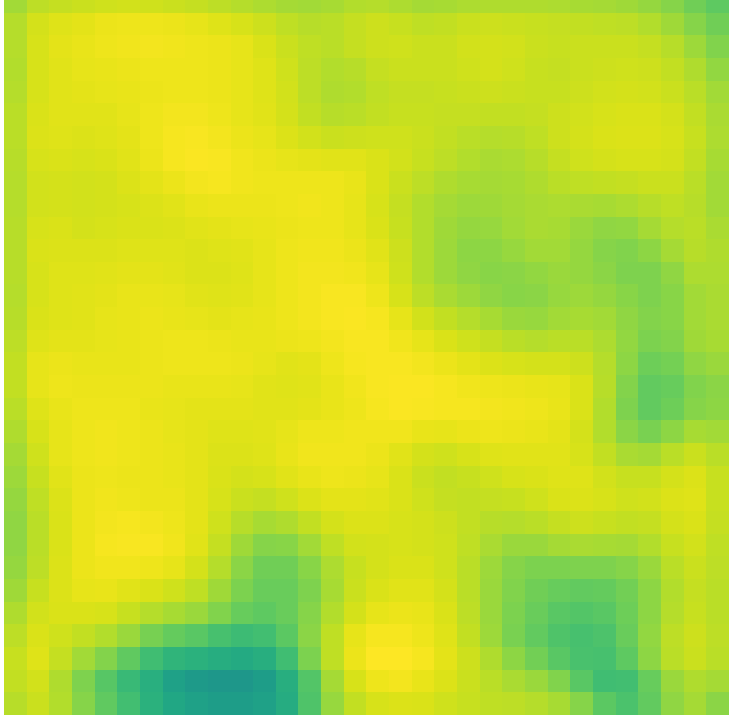}} 
		\end{array}$
	\end{center}
	\caption{Qualitative comparison between reference and predicted thermal maps for three representative examples.}
	\label{fig:qualitative_examples}
\end{figure*}

\begin{figure*}[h]
	\begin{center}$
		\begin{array}{cc}
			\subfloat[]{\includegraphics[scale=.22]{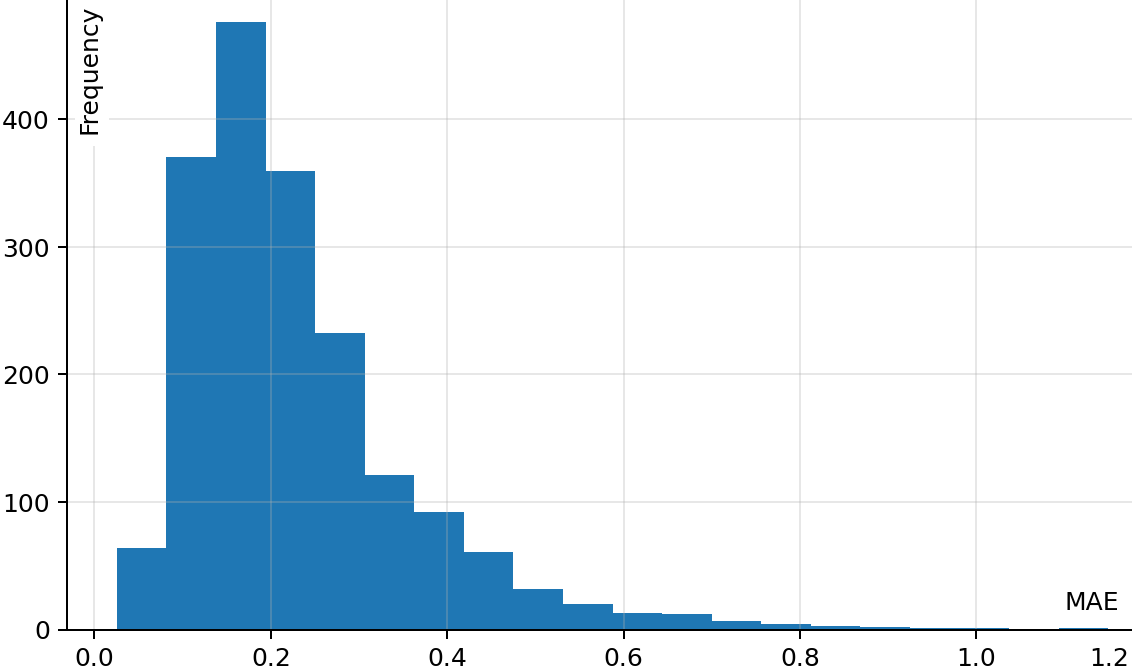}}&
			\subfloat[]{\includegraphics[scale=.2]{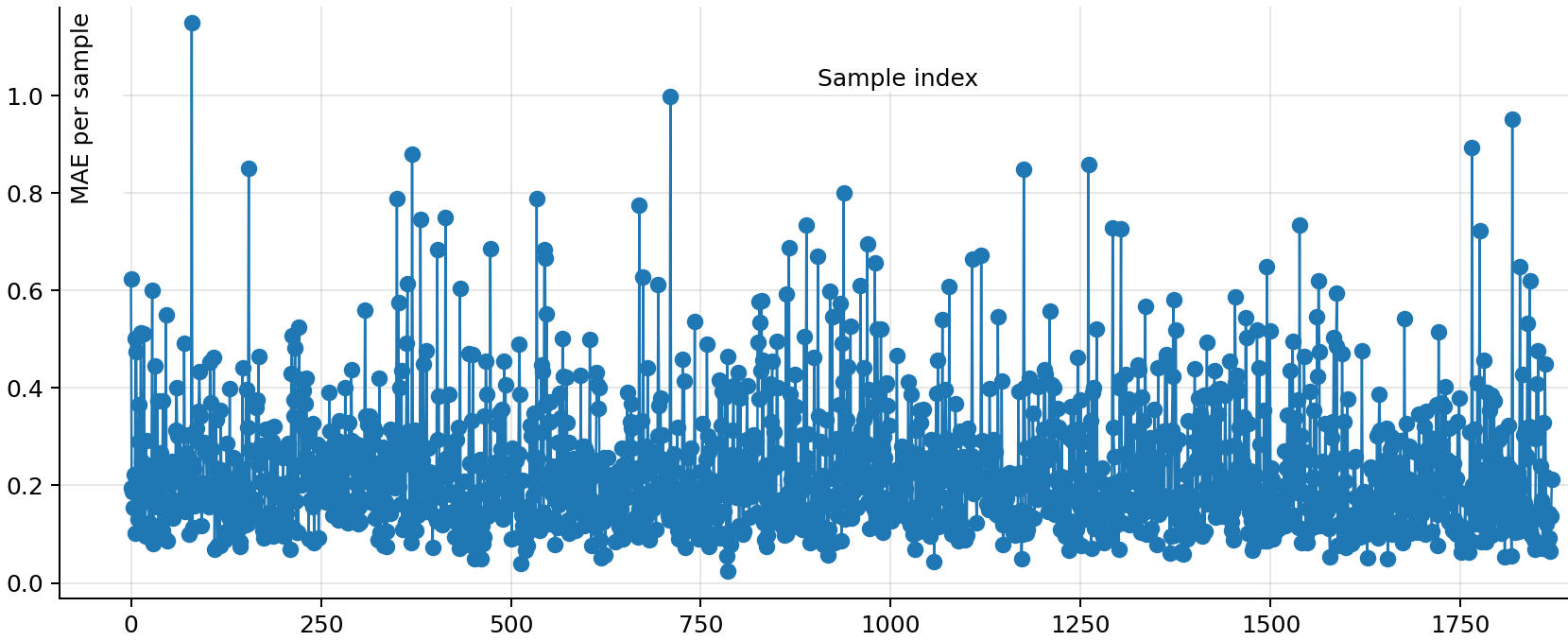}} 
		\end{array}$
	\end{center}
	\caption{(a) Distribution of per-sample MAE values. (b) Per-sample MAE across all evaluated samples.}
	\label{fig:mae-distribution}
\end{figure*}

\subsection{Qualitative Results}

Besides the global metrics, qualitative inspection of predicted maps was also performed. In general, the ConvLSTM model was able to recover the main large-scale thermal structure of the target maps. The warm regions and the main spatial gradients were usually well reconstructed, especially in representative medium-complexity cases. However, the predictions were still smoother than the reference MODIS maps. This means that the model was more successful in capturing coarse spatial patterns than very fine local anomalies. Such behavior is expected, since the target maps contain sharp local irregularities, while the model tries to learn stable spatiotemporal structure from limited and partially noisy observations. Figure~\ref{fig:qualitative_examples} presents three representative qualitative examples. The first row contains the reference thermal maps, while the second row shows the corresponding predictions produced by the ConvLSTM model. In the first two examples, the model follows the main spatial structure of the reference maps reasonably well and produces relatively small prediction errors. The corresponding errors for the predicted maps are as follows: first predicted map (MAE = 0.1548, RMSE = 0.2387) and second predicted map (MAE = 0.1319, RMSE = 0.1868). In the third example, however, the model fails to reproduce a more difficult and strongly localized thermal pattern, which results in substantially larger error values (MAE = 1.1496, RMSE = 1.1693). Overall, the qualitative analysis confirms that the model is generally able to capture the dominant thermal structure of the target maps, although the predicted maps remain smoother than the corresponding reference MODIS maps. To better understand the global behavior of the model over the full evaluation set, the per-sample MAE values were analyzed in two complementary ways. Figure~\ref{fig:mae-distribution}(a) shows the distribution of per-sample MAE values. It can be seen that most samples are concentrated in the low-to-moderate error range, while only a relatively small number of cases produce large errors. This indicates that the overall performance of the model is generally stable, although a limited number of difficult outliers still remain. Figure~\ref{fig:mae-distribution}(b) presents the MAE value for each evaluated sample. The plot shows that, for most samples, the error remains relatively low, while occasional peaks correspond to more challenging thermal patterns that are harder for the model to reproduce accurately.

\subsection{Discussion}

The experiments clearly indicate that dataset quality plays a major role in the final predictive performance. The initial experiments, which relied on a smaller and less informative dataset, produced smoother predictions and less stable quantitative results. After increasing the number of samples, extending the temporal coverage, and enriching the meteorological forcing with data from multiple Sarajevo locations, the overall performance improved substantially. The best results were achieved by the final ConvLSTM model trained with the hybrid loss function. Its $R^2$ score of $0.8877$ indicates a strong agreement between the predicted and reference thermal maps. At the same time, the qualitative analysis shows that some limitations still remain. In particular, the model tends to smooth out very fine local thermal details, and a limited number of challenging cases still produce relatively large errors. Nevertheless, the obtained results are strong enough to confirm the usefulness of the proposed practical framework. They show that a GPU-accelerated ConvLSTM model, combined with satellite-derived thermal observations and enriched daily meteorological forcing, can provide meaningful next-day urban thermal predictions for a real urban environment.

\section{Conclusion}\label{sec7}

This chapter presented a practical GPU-accelerated deep learning framework for next-day urban thermal prediction and simplified urban heat risk assessment. The proposed approach combined MODIS land surface temperature maps with daily meteorological forcing variables to model short-term thermal dynamics over Sarajevo. Special attention was given to multi-source dataset construction, spatiotemporal deep learning models, and the use of GPU acceleration to enable efficient repeated training and evaluation. The results showed that dataset design strongly affects predictive performance. While early experiments based on smaller datasets produced smoother and less stable predictions, substantial improvements were achieved after extending the temporal range, increasing the number of valid samples, and introducing multi-location meteorological forcing. The best performance was obtained with the ConvLSTM model trained using a hybrid loss function, reaching an MAE of 0.2293, an RMSE of 0.3089, and an $R^2$ score of 0.8877. These results indicate strong agreement between predicted and reference thermal maps, although some fine local details remain difficult to reconstruct. Overall, the findings confirm that GPU-accelerated deep learning provides a practical basis for urban thermal prediction in real environments. The proposed framework is flexible enough to support further methodological improvements and application-oriented extensions related to urban climate adaptation, early warning, and heat risk assessment.

\bibliography{sn-bibliography}

@article{doi:10.1371/journal.pone.0127277,
	author = {Morabito, M. and Crisci, A. and Gioli, B. and Gualtieri, G. and Toscano, P. and {Di Stefano}, V. and Orlandini, S. and Gensini, G. F.},
	title			= {Urban-hazard risk analysis: mapping of heat-related risks in the elderly in major Italian cities},
	journal		= {PLoS One},
	volume		= {18},
	number		= {5},
	year			= {2015},
	doi={10.1371/journal.pone.0127277}
}

@article{ZHANG2019852,
	title = {Mapping heat-related health risks of elderly citizens in mountainous area: A case study of Chongqing, China},
	journal = {Science of The Total Environment},
	volume = {663},
	pages = {852-866},
	year = {2019},
	issn = {0048-9697},
	doi = {https://doi.org/10.1016/j.scitotenv.2019.01.240},
	author = {Wei Zhang and Caigui Zheng and Feng Chen}
}

@article{doi:10.1177/23998083241280746,
	author = {Nicola Colaninno and Rounaq Basu and Maryam Hosseini and Abdulaziz Alhassan and Liu Liu and Andres Sevtsuk},
	title ={A sidewalk-level urban heat risk assessment framework using pedestrian mobility and urban microclimate modeling},
	journal = {Environment and Planning B: Urban Analytics and City Science},
	volume = {52},
	number = {5},
	pages = {1071-1090},
	year = {2025},
	doi = {10.1177/23998083241280746},
	URL = {https://doi.org/10.1177/23998083241280746},
	eprint = {https://doi.org/10.1177/23998083241280746}	
}

@article{Ge2025,
	author  = {Ge, Shuang and Zhan, Wenfeng and Li, Jiufeng and Li, Long and Dong, Pan and Li, Xiang and Wang, Chenguang and Wang, Chunli and Gao, Yihan},
	title   = {Prediction of next-day 1-km canopy urban heat island by integrating a multi-block convolutional neural network with satellite- and ground-based observations},
	journal = {Sustainable Cities and Society},
	volume  = {131},
	pages   = {106701},
	year    = {2025},
	doi     = {10.1016/j.scs.2025.106701}
}

@article{Lyu2022,
	author  = {Fangzheng Lyu and Shaohua Wang and Su Yeon Han and Charlie Catlett and Shaowen Wang},
	title   = {An integrated cyberGIS and machine learning framework for fine-scale prediction of Urban Heat Island using satellite remote sensing and urban sensor network data},
	journal = {Urban Informatics},
	volume  = {1},
	pages   = {6},
	year    = {2022},
	doi     = {10.1007/s44212-022-00002-4}
}

@Article{atmos14020343,
	AUTHOR = {Emmanuel, Rohinton and Jalal, Mushfik and Ogunfuyi, Samson and Maharoof, Nusrath and Zala, Megi and Perera, Narein and Ratnayake, Rangajeewa},
	TITLE = {Urban Heat Risk: Protocols for Mapping and Implications for Colombo, Sri Lanka},
	JOURNAL = {Atmosphere},
	VOLUME = {14},
	YEAR = {2023},
	NUMBER = {2},
	ARTICLE-NUMBER = {343},
	URL = {https://www.mdpi.com/2073-4433/14/2/343},
	ISSN = {2073-4433},
	DOI = {10.3390/atmos14020343}
}

@article{Kalfarisi2022,
	author  = {Kalfarisi, Rony and Chew, Alvin Wei Ze and Cai, Jianping and Xue, Meng and Pok, Jocelyn and Wu, Zheng Yi},
	title   = {Predictive modeling framework accelerated by GPU computing for smart water grid data-driven analysis in near real-time},
	journal = {Advances in Engineering Software},
	volume  = {173},
	pages   = {103287},
	year    = {2022},
	doi     = {10.1016/j.advengsoft.2022.103287}
}

@Article{su15031820,
	AUTHOR = {Técher, Magalie and Ait Haddou, Hassan and Aguejdad, Rahim},
	TITLE = {Urban Heat Island’s Vulnerability Assessment by Integrating Urban Planning Policies: A Case Study of Montpellier Méditerranée Metropolitan Area, France},
	JOURNAL = {Sustainability},
	VOLUME = {15},
	YEAR = {2023},
	NUMBER = {3},
	ARTICLE-NUMBER = {1820},
	URL = {https://www.mdpi.com/2071-1050/15/3/1820},
	ISSN = {2071-1050},
	DOI = {10.3390/su15031820}
}

@Article{rs16163032,
	AUTHOR = {Li, Fei and Yigitcanlar, Tan and Nepal, Madhav and Thanh, Kien Nguyen and Dur, Fatih},
	TITLE = {A Novel Urban Heat Vulnerability Analysis: Integrating Machine Learning and Remote Sensing for Enhanced Insights},
	JOURNAL = {Remote Sensing},
	VOLUME = {16},
	YEAR = {2024},
	NUMBER = {16},
	ARTICLE-NUMBER = {3032},
	URL = {https://www.mdpi.com/2072-4292/16/16/3032},
	ISSN = {2072-4292},
	DOI = {10.3390/rs16163032}
}

@ARTICLE{10.3389/fpubh.2022.989963,
	AUTHOR={Wu, Chaowei  and Shui, Wei  and Huang, Zhigang  and Wang, Chunhui  and Wu, Yuehui  and Wu, Yinpan  and Xue, Chengzhi  and Huang, Yunhui  and Zhang, Yiyi  and Zheng, Dongyang },	
	TITLE={Urban heat vulnerability: A dynamic assessment using multi-source data in coastal metropolis of Southeast China},
	JOURNAL={Frontiers in Public Health},
	VOLUME={Volume 10 - 2022},
	YEAR={2022},
	DOI={10.3389/fpubh.2022.989963},
	ISSN={2296-2565}
	}

@article{Wang2025,
	author  = {Lin Wang},
	title   = {Urban heat island monitoring and intelligent prediction method based on big data analysis},
	journal = {Sustainable Energy Research},
	volume={12},
	number={1},
	year    = {2025},
	doi={10.1186/s40807-025-00206-7}
}

@article{LiWang2021,
	author  = {Xiaojiang Li and Guoqing Wang},
	title   = {GPU parallel computing for mapping urban outdoor heat exposure},
	journal = {Theoretical and Applied Climatology},
	volume  = {145},
	number={3},
	pages   = {1101--1111},
	year    = {2021},
	doi={10.1007/s00704-021-03692-z}
}

@ARTICLE{10938603,
	author={Tang, Ning and Farhan, Muhammad and Mohammad, Pir and Abdullah-Al-Wadud, M. and Hussain, Saddam and Hamza, Umair and Zulqarnain, Rana Muhammad and Rebouh, Nazih Yacer},
	journal={IEEE Journal of Selected Topics in Applied Earth Observations and Remote Sensing}, 
	title={Enhancing Urban Heat Island Analysis Through Multisensor Data Fusion and GRU-Based Deep Learning Approaches for Climate Modeling}, 
	year={2025},
	volume={18},
	number={},
	pages={9279-9296},
	doi={10.1109/JSTARS.2025.3554529}
	}

@article{Shafiq2025,
	author  = {Fatima Shafiq and Amna Zafar and Muhammad Usman Ghani Khan and Sajid Iqbal and Abdulmohsen Saud Albesher and Muhammad Nabeel Asghar},
	title   = {Extreme heat prediction through deep learning and explainable AI},
	volume={20},
	number={3},
	journal = {PLOS One},
	year    = {2025},
	doi     = {10.1371/journal.pone.0316367}
}

@Article{s24020514,
	AUTHOR = {Mwitta, Canicius and Rains, Glen C. and Prostko, Eric},
	TITLE = {Evaluation of Inference Performance of Deep Learning Models for Real-Time Weed Detection in an Embedded Computer},
	JOURNAL = {Sensors},
	VOLUME = {24},
	YEAR = {2024},
	NUMBER = {2},
	ARTICLE-NUMBER = {514},
	URL = {https://www.mdpi.com/1424-8220/24/2/514},
	PubMedID = {38257609},
	ISSN = {1424-8220},
	DOI = {10.3390/s24020514}
}

@article{Pandey2022,
	author  = {Mohit Pandey and Michael Fernandez and Francesco Gentile and Olexandr Isayev and Alexander Tropsha and Abraham C. Stern and Artem Cherkasov},
	title   = {The transformational role of GPU computing and deep learning in drug discovery},
	journal = {Nature Machine Intelligence},
	volume  = {4},
	number={3},
	pages   = {211--221},
	year    = {2022},
	doi     = {https://doi.org/10.1038/s42256-022-00463-x}
}

@article{Lloyd2017,
	author  = {Christopher T. Lloyd and Alessandro Sorichetta and Andrew J. Tatem},
	title   = {High resolution global gridded data for use in population studies},
	journal = {Scientific Data},
	volume  = {4},
	number={1},
	pages   = {170001},
	year    = {2017},
	doi     = {https://doi.org/10.1038/sdata.2017.1}
}

@article{WAN200859,
	title = {New refinements and validation of the MODIS Land-Surface Temperature/Emissivity products},
	journal = {Remote Sensing of Environment},
	volume = {112},
	number = {1},
	pages = {59-74},
	year = {2008},
	issn = {0034-4257},
	doi = {https://doi.org/10.1016/j.rse.2006.06.026},
	url = {https://www.sciencedirect.com/science/article/pii/S0034425707003665},
	author = {Zhengming Wan}
}

@article{CHEVAL2024100603,
	title = {A systematic review of urban heat island and heat waves research (1991–2022)},
	journal = {Climate Risk Management},
	volume = {44},
	pages = {100603},
	year = {2024},
	issn = {2212-0963},
	doi = {https://doi.org/10.1016/j.crm.2024.100603},
	url = {https://www.sciencedirect.com/science/article/pii/S2212096324000202},
	author = {Sorin Cheval and Vlad-Alexandru Amihăesei and Zenaida Chitu and Alexandru Dumitrescu and Vladut Falcescu and Adrian Irașoc and Dana Magdalena Micu and Eugen Mihulet and Irina Ontel and Monica-Gabriela Paraschiv and Nicu Constantin Tudose}
}

@article{CUERDOVILCHES2023164412,
	title = {Impact of urban heat islands on morbidity and mortality in heat waves: Observational time series analysis of Spain's five cities},
	journal = {Science of The Total Environment},
	volume = {890},
	pages = {164412},
	year = {2023},
	issn = {0048-9697},
	doi = {https://doi.org/10.1016/j.scitotenv.2023.164412},
	url = {https://www.sciencedirect.com/science/article/pii/S0048969723030334},
	author = {T. Cuerdo-Vilches and J. Díaz and J.A. López-Bueno and M.Y. Luna and M.A. Navas and I.J. Mirón and C. Linares}
}

@article{Huang2023,
	author= {Wan Ting Katty Huang and Pierre Masselot and Elie Bou-Zeid and Simone Fatichi and Athanasios Paschalis and Ting Sun and Antonio Gasparrini and Gabriele Manoli},
	title = {Economic valuation of temperature-related mortality attributed to urban heat islands in European cities},
	journal = {Nature Communications},
	volume={14},
	doi={https://doi.org/10.1038/s41467-023-43135-z},
	number={1},
	pages={1--12},
	year= {2023}
}

@article{Hsu2021,
	author  = {Angel Hsu and Glenn Sheriff and Tirthankar Chakraborty and Diego Manya},
	title   = {Disproportionate exposure to urban heat island intensity across major US cities},
	journal = {Nature Communications},
	year    = {2021},
	pages={1--11},
	volume={12},
	number={1},
	doi={https://doi.org/10.1038/s41467-021-22799-5}
}

@article{Geophysics2023,
	author = {Barriopedro, D. and García-Herrera, R. and Ordóñez, C. and Miralles, D. G. and Salcedo-Sanz, S.},
	title = {Heat Waves: Physical Understanding and Scientific Challenges},
	journal = {Reviews of Geophysics},
	volume = {61},
	number = {2},
	pages = {e2022RG000780},
	doi = {https://doi.org/10.1029/2022RG000780},
	year = {2023}
}

@Article{atmos16010082,
	AUTHOR = {Zhang, Huijun and Liu, Yaxin and Zhang, Chongyu and Li, Ningyun},
	TITLE = {Machine Learning Methods for Weather Forecasting: A Survey},
	JOURNAL = {Atmosphere},
	VOLUME = {16},
	YEAR = {2025},
	NUMBER = {1},
	ARTICLE-NUMBER = {82},
	URL = {https://www.mdpi.com/2073-4433/16/1/82},
	ISSN = {2073-4433},
	DOI = {10.3390/atmos16010082}
}

@Article{su17083747,
	AUTHOR = {Xu, Chang and Wei, Ruihan and Tong, Hui},
	TITLE = {A Systematic Review of Methodological Advances in Urban Heatwave Risk Assessment: Integrating Multi-Source Data and Hybrid Weighting Methods},
	JOURNAL = {Sustainability},
	VOLUME = {17},
	YEAR = {2025},
	NUMBER = {8},
	ARTICLE-NUMBER = {3747},
	URL = {https://www.mdpi.com/2071-1050/17/8/3747},
	ISSN = {2071-1050},
	DOI = {10.3390/su17083747}
}

@article{BOUDREAULT2025109965,
	title = {Machine learning for modelling the health impacts of extreme heat: A comprehensive literature review},
	journal = {Environment International},
	volume = {206},
	pages = {109965},
	year = {2025},
	issn = {0160-4120},
	doi = {https://doi.org/10.1016/j.envint.2025.109965},
	url = {https://www.sciencedirect.com/science/article/pii/S0160412025007160},
	author = {Jérémie Boudreault and Félix Lamothe and Céline Campagna and Fateh Chebana}
}

@article{DUAN201916,
	title = {Validation of Collection 6 MODIS land surface temperature product using in situ measurements},
	journal = {Remote Sensing of Environment},
	volume = {225},
	pages = {16-29},
	year = {2019},
	issn = {0034-4257},
	doi = {https://doi.org/10.1016/j.rse.2019.02.020},
	url = {https://www.sciencedirect.com/science/article/pii/S0034425719300756},
	author = {Si-Bo Duan and Zhao-Liang Li and Hua Li and Frank-M. Göttsche and Hua Wu and Wei Zhao and Pei Leng and Xia Zhang and César Coll}
}

@article{Tatem2017,
	author  = {Andrew J. Tatem},
	title   = {WorldPop, open data for spatial demography},
	journal = {Scientific Data},
	year    = {2017},
	volume={4},
	pages={1--4},
	doi={https://doi.org/10.1038/sdata.2017.4}
}

@Article{hydrology11080127,
	AUTHOR = {Niño Medina, Johann Santiago and Suarez Barón, Marcó Javier and Reyes Suarez, José Antonio},
	TITLE = {Application of Deep Learning for the Analysis of the Spatiotemporal Prediction of Monthly Total Precipitation in the Boyacá Department, Colombia},
	JOURNAL = {Hydrology},
	VOLUME = {11},
	YEAR = {2024},
	NUMBER = {8},
	ARTICLE-NUMBER = {127},
	URL = {https://www.mdpi.com/2306-5338/11/8/127},
	ISSN = {2306-5338},
	DOI = {10.3390/hydrology11080127}
}

@Article{ijgi4042306,
	AUTHOR = {Shekhar, Shashi and Jiang, Zhe and Ali, Reem Y. and Eftelioglu, Emre and Tang, Xun and Gunturi, Venkata M. V. and Zhou, Xun},
	TITLE = {Spatiotemporal Data Mining: A Computational Perspective},
	JOURNAL = {ISPRS International Journal of Geo-Information},
	VOLUME = {4},
	YEAR = {2015},
	NUMBER = {4},
	PAGES = {2306--2338},
	URL = {https://www.mdpi.com/2220-9964/4/4/2306},
	ISSN = {2220-9964},
	DOI = {10.3390/ijgi4042306}
}

@InProceedings{10.1007/978-981-19-1122-4-47,
	author="Tuba, Eva
	and Tuba, Ira
	and Hrosik, Romana Capor
	and Alihodzic, Adis
	and Tuba, Milan",
	editor="Rathore, Vijay Singh
	and Sharma, Subhash Chander
	and Tavares, Joao Manuel R.S.
	and Moreira, Catarina
	and Surendiran, B.",
	title="Image Classification by Optimized Convolution Neural Networks",
	booktitle="Rising Threats in Expert Applications and Solutions",
	year="2022",
	publisher="Springer Nature Singapore",
	address="Singapore",
	pages="447--454",
	isbn="978-981-19-1122-4"
}

@Article{ijerph15040640,
	AUTHOR = {Voelkel, Jackson and Hellman, Dana and Sakuma, Ryu and Shandas, Vivek},
	TITLE = {Assessing Vulnerability to Urban Heat: A Study of Disproportionate Heat Exposure and Access to Refuge by Socio-Demographic Status in Portland, Oregon},
	JOURNAL = {International Journal of Environmental Research and Public Health},
	VOLUME = {15},
	YEAR = {2018},
	NUMBER = {4},
	ARTICLE-NUMBER = {640},
	URL = {https://www.mdpi.com/1660-4601/15/4/640},
	PubMedID = {29601546},
	ISSN = {1660-4601},
	DOI = {10.3390/ijerph15040640}
}

@article{Pan2024,
	author  = {Xiyu Pan and Dimitris Mavrokapnidis and Hoang T. Ly and Neda Mohammadi and John E. Taylor},
	title   = {Assessing and forecasting collective urban heat exposure with smart city digital twins},
	journal = {Scientific Reports},
	year    = {2024},
	doi={https://doi.org/10.1038/s41598-024-59228-8}, 
	volume={14},
	number={1},
	pages={1--14}
}

@article{DAmbrosio2023,
	author  = {Valeria D’Ambrosio and Ferdinando Di Martino and Vittorio Miraglia},
	title   = {A GIS-based framework to assess heatwave vulnerability and impact scenarios in urban systems},
	journal = {Scientific Reports},
	year    = {2023},
	page={1--18},
	volume={13},
	number={1},
	doi={https://doi.org/10.1038/s41598-023-39820-0}
}

@article{YE2025100870,
	title = {Billions of people exposed to increasing heat but decreasing greenness from 2000 to 2022},
	journal = {The Innovation},
	volume = {6},
	number = {5},
	pages = {100870},
	year = {2025},
	issn = {2666-6758},
	doi = {https://doi.org/10.1016/j.xinn.2025.100870},
	url = {https://www.sciencedirect.com/science/article/pii/S2666675825000736},
	author = {Tingting Ye and Rongbin Xu and Wenzhong Huang and Zhengyu Yang and Pei Yu and Wenhua Yu and Yanming Liu and Yao Wu and Bo Wen and Yiwen Zhang and Jaime E. Hart and Mark Nieuwenhuijsen and Michael J. Abramson and Yuming Guo and Shanshan Li}
}

\end{document}